\documentclass[lettersize,journal]{IEEEtran}
\usepackage{amsmath,amsfonts}
\usepackage{algorithmic}
\usepackage{array}
\usepackage[caption=false,font=normalsize,labelfont=sf,textfont=sf]{subfig}
\usepackage{textcomp}
\usepackage{stfloats}
\usepackage{url}
\usepackage{verbatim}
\usepackage{graphicx}
\usepackage{multirow}
\usepackage{color,xcolor}
\usepackage{soul}
\usepackage{colortbl}
\usepackage{float}
\usepackage{subfig}
\usepackage{booktabs}
\usepackage{fontawesome}
\usepackage{hyperref}
\usepackage{makecell}

\let\hl\relax

\def\BibTeX{{\rm B\kern-.05em{\sc i\kern-.025em b}\kern-.08em
    T\kern-.1667em\lower.7ex\hbox{E}\kern-.125emX}}
\usepackage{balance}
\begin{document}
\title{Learning Suspected Anomalies from Event Prompts for Video Anomaly Detection}
\author{Chenchen Tao\IEEEauthorrefmark{1}, Xiaohao Peng\IEEEauthorrefmark{1}, Chong Wang\textsuperscript{\faEnvelope}, ~\IEEEmembership{Member,~IEEE},  Jiafei Wu, Puning Zhao, Jun Wang, Jiangbo Qian
\thanks{




\IEEEauthorrefmark{1} These authors contributed equally to this work and should be considered co-first authors.

\faEnvelope \ Corresponding Author: Chong Wang.}
}

\markboth{Journal of \LaTeX\ Class Files,~Vol.~18, No.~9, September~2020}%
{Learning Suspected Anomalies from Event Prompts for Video Anomaly Detection}

\maketitle

\begin{abstract}
Most models for weakly supervised video anomaly detection (WS-VAD) rely on multiple instance learning, aiming to distinguish normal and abnormal snippets without specifying the type of anomaly. However, the ambiguous nature of anomaly definitions across contexts may introduce inaccuracy in discriminating abnormal and normal events. To show the model what is anomalous, a novel framework is proposed to guide the learning of suspected anomalies from event prompts. Given a textual prompt dictionary of potential anomaly events and the captions generated from anomaly videos, the semantic anomaly similarity between them could be calculated to identify the suspected events for each video snippet. It enables a new multi-prompt learning process to constrain the visual-semantic features across all videos, as well as provides a new way to label pseudo anomalies for self-training.
To demonstrate its effectiveness, comprehensive experiments and detailed ablation studies are conducted on four datasets, namely XD-Violence, UCF-Crime, TAD, and ShanghaiTech. Our proposed model outperforms most state-of-the-art methods in terms of AP or AUC (86.5\%, \hl{90.4}\%, 94.4\%, and 97.4\%). Furthermore, it shows promising performance in open-set and cross-dataset cases.
The data, code, and models can be found at: \url{https://github.com/shiwoaz/lap}.
\end{abstract}


\section{Introduction}
\IEEEPARstart{V}{ideo} anomaly detection (VAD) \cite{zhao2017spatio,yu2020cloze, tao2024feature} is crucial in video surveillance, given the extensive use of surveillance cameras. The task of VAD is to determine whether each frame in a video is normal or abnormal, which poses a significant challenge as it is not feasible to train a model with complete supervision. Consequently, weakly supervised learning methods (WS-VAD) \cite{yu2021cross, shi2023abnormal,acmmm1} that solely rely on video-level annotations have gained importance and popularity in recent years.

\begin{figure}
    \centering
    \includegraphics[width=1\linewidth]{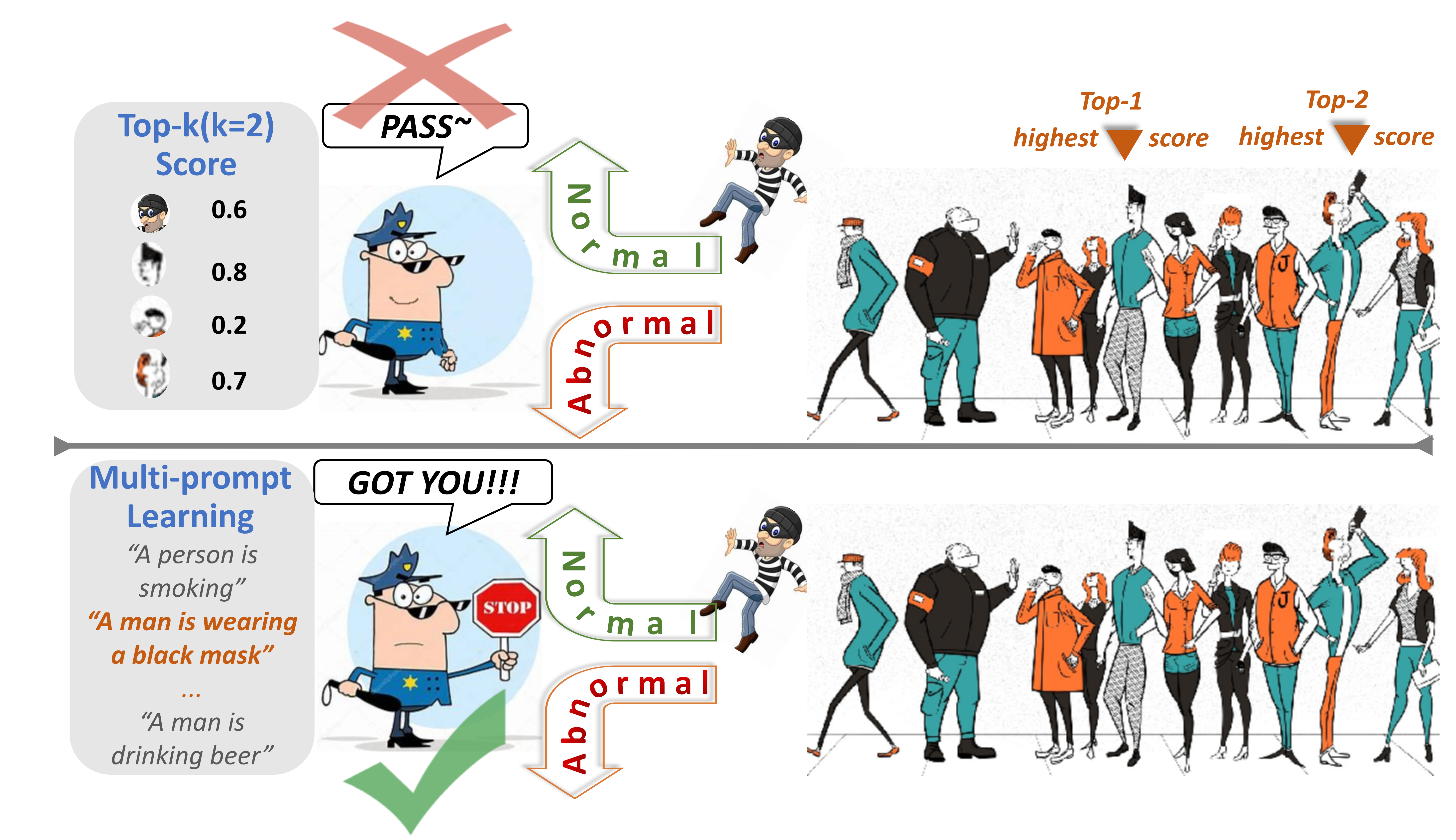}
    \caption{The difference between the traditional multiple instance learning methods (upper) and our model (lower). The former one only learns the anomalies using top-$k$ scores in each abnormal video, while the latter utilizes a prompt dictionary to provide extra guidance across different videos.}
    \label{fig:abstract_fig}
\end{figure}

The general paradigm of these methods involves utilizing convolutional networks such as 3D ConvNet (C3D) \cite{tran2015learning}, inflated 3D ConvNet (I3D) \cite{carreira2017quo}, or vision transformer \cite{dosovitskiy2020image} to extract visual features and aggregate spatio-temporal information between consecutive frames. Subsequently, an anomaly-detection network is trained using multiple instance learning (MIL) \cite{sultani2018real}. This approach simultaneously maximizes and minimizes the top-$k$ highest scores from individual anomaly and normal videos, respectively. Most methods \cite{sultani2018real, tian2021weakly} only focused on the visual-related modality, while some \cite{chen2023tevad,wu2023vadclip} have incorporated semantic descriptions into videos. However, such semantic information was simply fused with the visual one, instead of delving into the underlying meaning of the textual descriptions. As a result, the MIL based approaches often suffer from a relatively high false alarm rate (FAR) and low accuracy in detecting ambiguous abnormal events.

Meanwhile, foundation models in natural language processing (NLP) and computer vision (CV), such as InstructGPT \cite{ouyang2022training} and CLIP \cite{radford2021learning}, have demonstrated impressive performance on multimodal tasks. Additionally, prompting techniques in the image field provide a new way to transfer semantic information from well-trained foundation models into vision tasks. It is intriguing to explore whether CLIP's zero-shot detection ability can be effectively transferred into video anomaly detection.


Therefore, a novel framework to \hl{\textbf{L}earn suspected \textbf{A}nomalies from event \textbf{P}rompts, called LAP}, is proposed in this paper. As illustrated in Figure \ref{fig:abstract_fig}, a prompt dictionary is designed to list the potential anomaly events. In order to mark suspected anomalies, it is utilized to compare with the captions generated from anomaly videos in the form of semantic features. As a result, an anomaly vector that records the most suspected anomalous events for each video snippet can be obtained. This vector is used to guide a new multi-prompt learning scheme across different videos, as well as form a new set of pseudo anomaly labels.



The main contributions of this work are threefold:

\begin{itemize}

\item The new textual prompts describing the abnormal events are introduced into weakly supervised video anomaly detection. Giving the explanation of what is anomalous, the score predictor can implicitly learn more details about the anomalies. It leads to incredible performance on open-set and cross-database problems.

\item A new multi-prompt learning strategy is proposed to provide an overall understanding of normal and abnormal patterns across different videos, while MIL is limited to individual videos.

\item Additional pseudo labels are excavated from the anomaly videos according to the semantic similarity between the event prompts and videos. They are utilized to train the predictor effectively in a self-supervised manner. 

\end{itemize}

\section{Related Works}
\label{sec:Related Works}

\subsection{Weakly Supervised Video Anomaly Detection}
The weakly supervised methods tackle frame-level anomaly detection by video-level annotations. Most of them are based primarily on multiple instance learning (MIL) due to limited annotated labels \cite{sultani2018real}. 
However, conventional MIL faces challenges in providing sufficient supervision for various anomalies, leading to misclassifications and a high false alarm rate. To address these issues, Yu et al. propose cross-epoch learning (XEL) \cite{yu2021cross}, which stores hard instances from previous epochs to optimize the anomaly predictor in the latest epoch. Additionally, dual memory units with uncertainty regulation (UR-DMU) \cite{zhou2023dual} extend the anomaly memory unit into learnable dual memory units to alleviate the high false alarm rate issue. Another approach called robust temporal feature magnitude learning (RTFM) \cite{tian2021weakly} trains a feature magnitude learning function to effectively recognize positive instances. All these methods are based on single or multiple visual modalities, including RGB and optical flow.

As the field embraces multi-modality models like GPT \cite{ouyang2022training} and CLIP \cite{radford2021learning}, researchers are now focusing on text-visual models. Text-empowered video anomaly detection (TEVAD) \cite{chen2023tevad} demonstrates improvements by generating text and visual features independently. However, TEVAD treats text features as auxiliary to visual features. In contrast, our approach aims to capitalize on the high semantic-level guidance provided by text, offering a unique perspective for enhancing anomaly detection performance.

\subsection{Prompt Tuning for Visual Tasks}
In the realm of pre-trained foundation multimodality models, a cost-effective prompt tuning approach is gaining traction for adapting models to downstream tasks in the domains of natural language processing (NLP) \cite{li2021prefix, lester2021power} and computer vision \cite{jia2022visual}.

The concept of prompt tuning originated in computer vision to tackle zero-shot or few-shot image tasks by incorporating semantic guidance. Multimodal models such as CLIP \cite{radford2021learning} leverage textual prompts for image classification, demonstrating state-of-the-art performance. In the video domain, Sato \cite{sato2023prompt} explores prompt tuning for zero-shot anomaly action recognition, using skeleton features and text embeddings in a shared space to refine decision boundaries. Wang et al. introduce prompt learning for action recognition (PLAR) \cite{wang2023prompt}, which incorporates optical flow and learnable prompts to acquire input-invariant knowledge from a prompt expert dictionary and input-specific knowledge based on the data.

A previous effort, the prompt-based feature mapping framework (PFMF) \cite{liu2023generating}, applies prompt-based learning to semi-supervised video anomaly detection. PFMF generates anomaly prompts by concatenating anomaly vectors from virtual datasets and scene vectors from real datasets, guiding the feature mapping network. However, the prompt in PFMF defines anomalies at the visual level, introducing ambiguity. In our work, we propose textual anomaly prompts based on prior knowledge to mine fine-grained anomalies to achieve high performance.

Several recent studies have introduced additional information or tasks to maximize the CLIP's effectiveness in WS-VAD. CLIP-assisted temporal self-attention (CLIP-TSA) \cite{joo2023clip} incorporates temporal information into CLIP features using a self-attention mechanism. In contrast, VadCLIP \cite{wu2023vadclip} delves deeper into aligning textual category labels with CLIP's visual features to enhance its WS-VAD performance. Unlike VadCLIP, which constructs learnable prompts based on class labels, our approach designs event prompts to describe specific anomaly-related situations, eliminating the need for additional supervised information or learning tasks.

\section{Methodology}
\begin{figure*}
    \centering
    \includegraphics[width=1\textwidth]{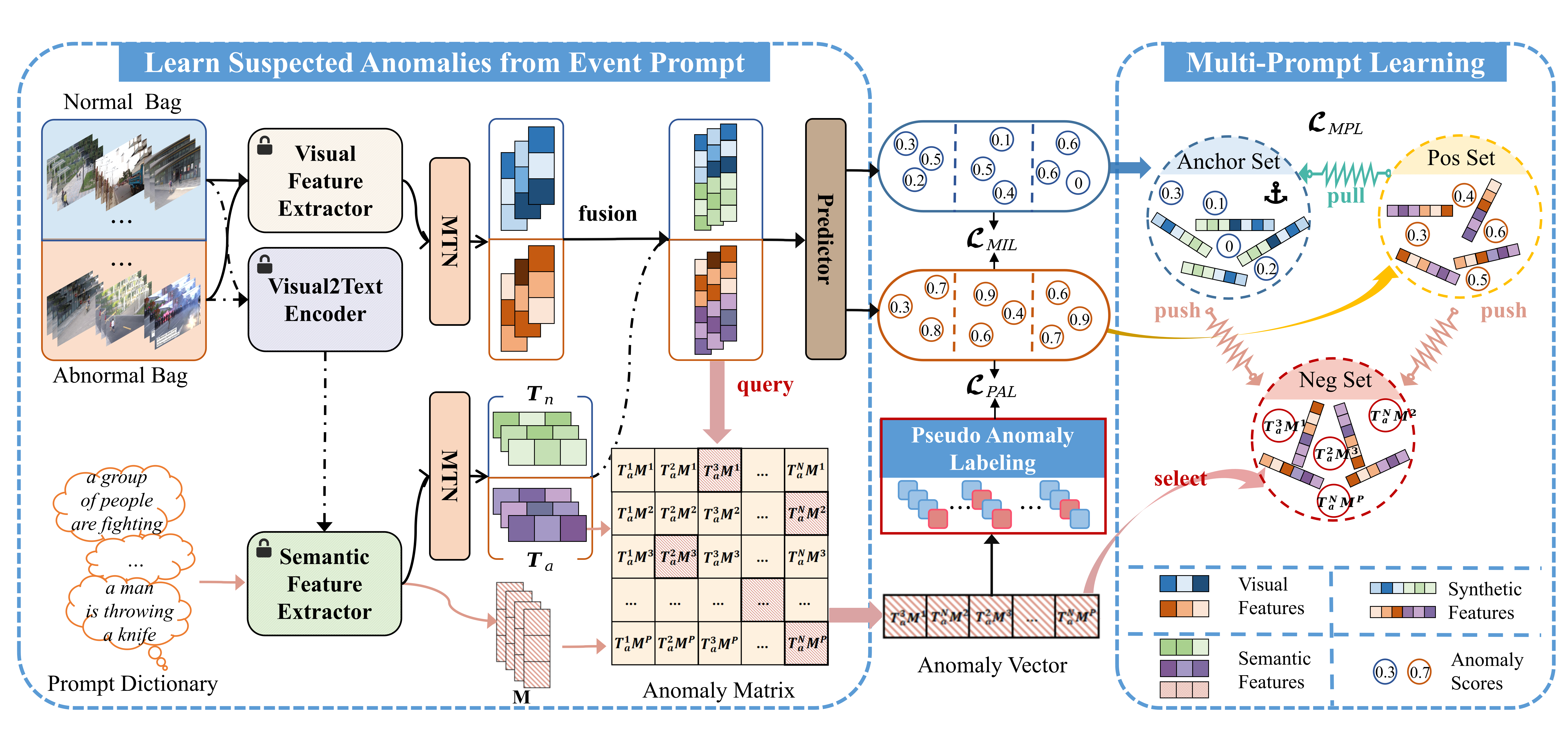}
    \caption{The overview of the proposed LAP framework. Synthetic features, as input to score predictors, are generated through the visual and semantic feature extractors. 
    A prompt dictionary is used to produce the anomaly matrix and vector, which is employed to perform multi-prompt learning (MPL) and pseudo anomaly labeling (PAL) across different videos.}
    \label{fig:main}
\end{figure*}
\label{sec:Proposed Method}
The proposed LAP framework, as shown in Figure \ref{fig:main}, is built upon the basic VAD structure consisting of a visual feature extractor and a score predictor.
To enhance discrimination between normal and abnormal videos, semantic clues from anomaly events are integrated using a prompt dictionary and an additional semantic feature extractor. This integration introduces three key processes: feature synthesis, multi-prompt learning, and pseudo anomaly labeling.
Semantic features are extracted from videos and fused with visual features, enriching the overall representation. Simultaneously, anomaly prompts, describing abnormal events, are employed to generate another set of semantic features. An anomaly similarity matrix is then computed between these two semantic feature sets. This matrix identifies the most anomalous features corresponding to each prompt in the dictionary. This batch-level anomaly vector not only facilitates a new multi-prompt learning procedure but also acts as a set of snippet-level pseudo labels. The subsequent subsections delve into the specifics of these procedures.

\subsection{Feature Synthesis}
Following the protocol of WS-VAD \cite{cao2023weakly}, we adopt a training approach using pairwise normal and abnormal data. Each training batch comprises an abnormal bag and a normal bag, consisting of $b$ abnormal and normal videos with labels $\textbf{y}_a=\textbf{1}\in \mathbb{R}^{b\times 1}$ and $\textbf{y}_n=\textbf{0}\in \mathbb{R}^{b\times 1}$, respectively. In this setup, every video is divided into $L$ snippets, each containing 16 consecutive frames. Consequently, the total number of snippets in each bag is $N = b \times L$. All of these snippets are then processed by the visual and semantic feature extractors.

To clarify, we acquire the visual features $\textbf{V}_a\in\mathbb{R}^{N\times d_v}$ and $\textbf{V}_n\in\mathbb{R}^{N\times d_v}$ from video snippets in the abnormal and normal bags, utilizing the visual encoder of a CLIP model \cite{radford2021learning}. Given that many VAD videos, primarily from surveillance, often lack associated text descriptions, we leverage a pre-trained visual-to-text encoder from SwinBERT \cite{lin2022swinbert}, following TEVAD \cite{chen2023tevad}, to generate descriptions for each video snippet. These textual descriptions then undergo processing by the semantic feature extractor (SimCSE \cite{gao2021simcse}), producing corresponding semantic features $\textbf{T}_a\in \mathbb{R}^{N\times d_t}$ and $\textbf{T}_n\in\mathbb{R}^{N\times d_t}$ for abnormal and normal video snippets. \hl{With extracted visual features and semantic features in hand, we feed these features into a multi-scale temporal network (MTN) to obtain both local and global temporal fused features.}

Intuitively, a combination of visual and semantic features is employed to synthesize new features $\textbf{F}_a\in\mathbb{R}^{N\times d_f}$ and $\textbf{F}_n\in\mathbb{R}^{N\times d_f}$, aiming for an enhanced feature representation,
\begin{align}
    \textbf{F}_a &= \theta(\textbf{V}_a, \textbf{T}_a),\\
    \textbf{F}_n &= \theta(\textbf{V}_n, \textbf{T}_n),
\end{align}
where $\theta$ symbolizes a feature alignment and fusion operation. \hl{It can be either a concatenation or addition.} Subsequently, the anomaly scores $\textbf{s}_a$ and $\textbf{s}_n$ can be calculated by applying $\textbf{F}_a$ and $\textbf{F}_n$ to a score predictor. Typically, this predictor takes the form of a multi-layer perceptron (MLP) \cite{popescu2009multilayer}, expressed as:
\begin{align}
\label{eqn:MLP}
   \textbf{s}_a &= \rm MLP(\textbf{F}_a),\\
   \textbf{s}_n &= \rm MLP(\textbf{F}_n).
\end{align}

\begin{figure*}
    \centering
    \includegraphics[width=1\textwidth]{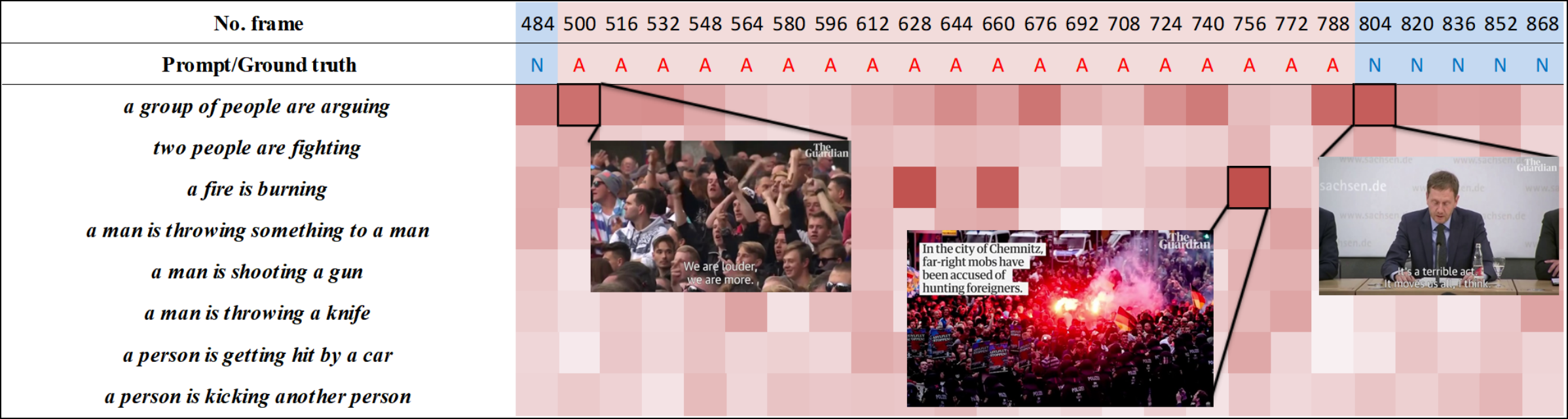}
    \caption{Visualization of the proposed anomaly matrix $\Psi^\top$. It is truncated due to the limited column width.}
    \label{fig:anomaly affinity matrix}
\end{figure*}

\subsection{Multi-Prompt Learning}
\label{sec:top prompt learning}
In recent WS-VAD, the prevalent approach for training the anomaly score predictor involves a multiple instance learning (MIL) framework \cite{chen2023tevad, lv2023unbiased}. This framework selects the top-$k$ highest anomaly scores from each video, whether abnormal or normal and employs their average $\hat{\textbf{y}}$ as the predicted value for the respective video. Given the complete score set $\textbf{s}= [\textbf{s}_a; \textbf{s}_n] \in\mathbb{R}^{2N\times 1}$,
\begin{equation}
    \begin{aligned}
\label{eqn:mil_vector}
   \hat{y} &= \frac{1}{k} \sum{\rm max}_k(\textbf{s}[i+1:i+L]), \\
    i &= (j-1)*L,\enspace j=1,2,...,2b,
\end{aligned}
\end{equation}
where max$_k(\textbf{s})$ denotes the operator to select $k$ largest values from vector $\textbf{s}$, $k$ is usually from 2 to 5, $i$ and $j$ indicate the snippet and video indices, respectively.
Then the MIL loss $\mathcal{L}_{\rm MIL}$ is formulated as,
\begin{align}
\label{eq:mil_loss}
    \mathcal{L}_{\rm MIL} = &{\sum_{y\in [\textbf{y}_a;\textbf{y}_n]} -(y\log(\hat{y})+(1-y)\log(1-\hat{y}))}.
\end{align}

From Equation \ref{eqn:mil_vector}, it can be seen that the top-$k$ strategy focuses only on a few snippets with the highest scores within an individual video. Moreover, the top anomaly scores in an abnormal video may not be from an abnormal snippet. Therefore, a new textual prompt dictionary consisting of $P$ anomaly prompts is designed to link abnormal video snippets from different videos. \hl{Unlike the category annotations used in VadCLIP {\cite{wu2023vadclip}}, we expanded the single word annotations into complete anomaly sentences, like ``someone is doing something to whom" or ``something is what". These sentences can better describe the events/actions related to a certain anomaly category. Then, the prompt dictionary is constructed as a set of these anomaly sentences,} e.g. ``A man is shooting a gun" or ``Something is on fire". As depicted in Figure \ref{fig:main}, these prompts undergo the same semantic extraction process (SimCSE \cite{gao2021simcse}) as the earlier video captions, generating their respective semantic features $\textbf{M}\in \mathbb{R}^{P \times d_t}$. Subsequently, we calculate the similarity between each prompt in the dictionary and every snippet in $\textbf{T}_a$ to construct an anomaly matrix $\Psi \in \mathbb{R}^{N \times P}$ as, 
\begin{align}
\label{eq:pseudo label}
    \Psi &= \frac{\textbf{T}_a \cdot \textbf{M}^\top}{\Vert \textbf{T}_a \Vert \Vert \textbf{M}\Vert},
\end{align}
in which $||\cdot||$ denotes the $l^2$-norm. The consideration of $\textbf{T}_n$ is dismissed here since there are no abnormal snippets in the normal bag. In essence, each element in $\Psi$ provides insights into the probable type of anomaly associated with each snippet or indicates where a predetermined abnormal event might occur. Figure \ref{fig:anomaly affinity matrix} offers a visual representation of $\Psi$, where frames containing abnormal events exhibit more pronounced colors. Notably, there is a discernible alignment between the frames and prompts.

In order to exploit the anomaly features across different videos, the most likely anomalous event of each snippet, i.e. the highest values in each row of $\Psi$, is picked to construct a new anomaly vector $\textbf{c}\in \mathbb{R}^{N \times 1}$.

To leverage these potential anomaly samples, we introduce a novel multi-prompt learning strategy. Based on the predicted score $\textbf{s}$ and the anomaly vector $\textbf{c}$, all features in $\textbf{F}_n$ and $\textbf{F}_a$ are categorized into three sets: anchor set, positive set, and negative set. Subsequently, their averages are computed, denoted as $f_{anc}$, $f_{pos}$, and $f_{neg}$. It's important to note that $f_{anc}$ and $f_{pos}$ model the normal features in normal and abnormal videos, respectively, and can be expressed as,
\begin{align}
f_{anc} &= \frac{1}{P}\sum_{i\in {\rm arg min}_P(\textbf{s}_n)} \textbf{F}_n[i,:],\\
     f_{pos} &= \frac{1}{P}\sum_{i\in {\rm argmin}_P(\textbf{s}_a)} \textbf{F}_a[i,:],      
\end{align}
where ${\rm arg min}_P(\textbf{s})$ denotes the operator to obtain the indices of $P$ lowest values in vector $\textbf{s}$, $\textbf{F}_n[i,:]$ and $\textbf{F}_a[i,:]$ are the $i$-th row of $\textbf{F}_n$ and $\textbf{F}_a$, respectively, which is a synthetic feature vector to represent a certain video snippet. In contrast, the negative set is built by choosing the most anomalous samples in anomaly videos, according to the similarity values in $\textbf{c}$. Thus the feature $f_{neg}$ can be formulated as,
\begin{align}
      f_{neg} &= \frac{1}{P}\sum_{i\in {\rm argmax}_P(\textbf{c})} \textbf{F}_a[i,:].
\end{align}
where ${\rm arg max}_P(\textbf{c})$ denotes the operator to obtain the indices of $P$ largest values in vector $\textbf{c}$.

Based on these representative features, it is possible to provide an overall understanding of normal and abnormal patterns across different videos. Thus, the multi-prompt learning loss $\mathcal{L}_{\rm MPL}$ is defined in a form of triplet loss, 
\begin{equation}
    \begin{aligned}
         \mathcal{L}_{\rm MPL} = \max (&\left\lVert f_{anc}
         -f_{pos}\right\rVert^2\\
         -&\left\lVert f_{anc}-f_{neg}\right\rVert^2+\alpha,0),
    \end{aligned}
\end{equation}
where $\alpha$ represents the margin coefficient. The goal of $\mathcal{L}_{\rm MPL}$ is to establish a considerable distance between  $f_{neg}$ and both $f_{anc}$ and $f_{pos}$ while simultaneously bringing $f_{anc}$ and $f_{pos}$ closer together. This feature-level examination implicitly impacts the training of the score predictor, given that the selection of $f_{anc}$ and $f_{pos}$ is based on $\textbf{s}$. 

\subsection{Pseudo Anomaly Labeling}
\label{sec:pseudo anomaly mining}

In addition to constructing the negative set in MPL, the anomaly vector $\textbf{c}$ serves as a metric for pseudo-labels, enabling the extraction of more latent information in the anomaly bag $\textbf{T}_a$. Specifically, the snippet-level pseudo-anomaly label $\textbf{p}$ is determined by a dynamic threshold within the current batch,
\begin{align}
        \textbf{p}[i] &=
\begin{cases}
1&, \text{if $\textbf{c}[i] > \mathcal{G}_{h}$}\\
0&, \text{otherwise}
\end{cases}, i=1,2,...,N,\\
\label{eqn:thred}
\quad\mathcal{G}_{h} &= {\rm mean}\{\textbf{c}\} + 
\tau \cdot {\rm std}\{\textbf{c}\},
    \end{align}
where $\textbf{p}[i]$ and $\textbf{c}[i]$ are the $i$-th element of $\textbf{p}$ and $\textbf{c}$, mean$\{\textbf{c}\}$ and std$\{\textbf{c}\}$ are the mean and standard deviation considering the anomaly vector $\textbf{c}$, and $\tau$ is a hyper-parameter.
Then, the anomaly score predictor can be trained in a fully supervised manner, through a pseudo anomaly loss $\mathcal{L}_{\rm PAL}$,
\begin{equation}
\begin{aligned}
    \mathcal{L}_{\rm PAL} = &\sum_{i=1}^{N}-(\textbf{p}[i]\log(\textbf{s}_a[i])\\
    &+(1-\textbf{p}[i])\log(1-\textbf{s}_a[i])).
\end{aligned}
\end{equation}

By incorporating prior knowledge into the pseudo label, the PAL module can better distinguish fine-grained anomalies and generate more accurate detecting results across abnormal videos.

The final training loss $\mathcal{L}_{\rm LAP}$ can be denoted as,
\begin{align}
    \mathcal{L}_{\rm LAP} = \mathcal{L}_{\rm MIL}+\beta \mathcal{L}_{\rm MPL} + \gamma \mathcal{L}_{\rm PAL},
\end{align}
where $\beta$ and $\gamma$ are hyper-parameters utilized in our model. Importantly, it's worth noting that the MPL and PAL modules are trained collaboratively. During the inference stage, the test samples will only traverse the feature extractors and the predictor to acquire abnormal scores, and the MPL and PAL modules incur no additional computational cost.

\subsection{Inference Process}
\hl{The inference process is identical to the baseline model, i.e. TEVAD {\cite{chen2023tevad}}, which is the left part of Figure {\ref{fig:main}} without the prompt dictionary. We initially extract visual and text features, which are processed through the feature alignment and fusion operation. Then, the fused features are fed into the anomaly predictor to calculate the anomaly score for each video snippet.
}

\section{Experiments}
\begin{table}[!t]
\footnotesize
\caption{Performance comparison of state-of-the-art methods on XD-Violence (AP\%) and UCF-Crime (AUC\%). \textbf{Bold} and \underline{underline} indicate the best and second-best results.}
\label{tab:xdsota}
\centering
\renewcommand\arraystretch{1.1}
\begin{tabular}{c|ccc|c|cc}
    \toprule
    \toprule
    \multirow{3}[1]{*}{\rotatebox{90}{Type}}&\multirow{3}[1]{*}{Source} &\multirow{3}[1]{*}{Method}&\multirow{3}[1]{*}{Feat.}&XD & \multicolumn{2}{c}{UCF}\\ \cmidrule{5-7}
    &  &&& \multirow{2}[1]{*}{AP}& AUC&AUC\\
    &  &&& & $_{all}$&$_{abn}$\\
    \midrule
    \midrule
    \multirow{2}[1]{*}{\rotatebox{90}{Semi}}&  CVPR 16'&Conv-AE \cite{hasan2016learning}&AE& 30.7 & -&-\\
    &  CVPR 22'&GCL \cite{zaheer2022generative}&CNN& -&71.0&- \\\midrule
    \multirow{13}[1]{*}{\rotatebox{90}{Weakly}} &   ICCV 21'&RTFM \cite{tian2021weakly}&CLIP & 78.3 & 85.7&63.9\\
    &  AAAI 22'&MSL \cite{li2022self} &ViT& 78.6& 85.6&-\\
    &  ECCV 22'&CSL-TAL \cite{panariello2022consistency} &I3D& 71.7&-&-\\
    &  CVPR 22'&BN-SVP \cite{sapkota2022bayesian} &I3D&-& 83.4&-\\
    &  CSVT 23'&Yang \cite{yang2023towards}&I3D & 77.7& 81.5&-\\
    & AAAI 23'&UR-DMU \cite{zhou2023dual} &CLIP&82.4&86.7&68.6\\
    &  CVPR 23'&ECUPL \cite{zhang2023exploiting} &I3D& 81.4& 86.2&-\\
    & CVPR 23'&CMRL \cite{cho2023look}&I3D &81.3 & 86.1&-\\
    &  CVPR 23'&TEVAD \cite{chen2023tevad}&I3D& 79.8& 84.9&-\\
    &  AAAI 23'&MGFN \cite{chen2023mgfn}&ViT& 80.1&-&-\\
    &  CVPR 23'&UMIL \cite{lv2023unbiased}&XCLIP &-& 86.7&68.7\\
    &  ICIP 23'&CLIP-TSA \cite{joo2023clip} &CLIP&82.2&87.6&-\\
    & CVPR 24'&Wu \textit{et al.} \cite{wu2023open}&CLIP&66.5&86.4&-\\
    & AAAI 24' &VadCLIP \cite{wu2023vadclip} &CLIP&\underline{84.5}&88.0 &70.2\\
    & \multirow{2}[1]{*}{\textbf{ours}} & \multirow{2}[1]{*}{\textbf{LAP}} & \textbf{CLIP} & \multirow{2}[1]{*}{\textbf{86.5}}&\multirow{2}[1]{*}{\underline{88.9}}&\multirow{2}[1]{*}{\underline{73.0}}\\
    & & & (\textit{SwinBert}) & & & \\
    & \multirow{2}[1]{*}{\textbf{ours}} & \multirow{2}[1]{*}{\textbf{LAP}} & \textbf{CLIP} &  \multirow{2}[1]{*}{-}&\multirow{2}[1]{*}{\textbf{90.4}}&\multirow{2}[1]{*}{\textbf{76.1}}\\
    & & & (\textit{UCA}) & & & \\
     \bottomrule
    \bottomrule
    \end{tabular}
\end{table}

In this section, the performance of our LAP model is evaluated on four datasets, namely XD-Violence \cite{wu2020not}, UCF-Crime \cite{sultani2018real}, TAD \cite{lv2021localizing} and ShanghaiTech \cite{zhong2019graph}. The area under the precision-recall curve, also known as the average precision (AP) is employed as the evaluation metric for XD-Violence following the protocol in \cite{zhang2023exploiting}. For UCF-Crime, TAD and ShanghaiTech, the area under the curve (AUC) of the frame-level receiver operating characteristics (ROC) is used instead. 
Specifically, AUC$_{all}$ represents the AUC for all testing videos, while AUC$_{abn}$ focuses only on abnormal videos in test set. The false alarm rates for all videos (FAR$_{all}$) and abnormal videos (FAR$_{abn}$) are also reported in our ablation studies.

\subsection{Datasets}
\textbf{XD-Violence} \cite{wu2020not} is a multi-scene public dataset for VAD. It consists of a total duration of 217 hours and includes 4,754 untrimmed videos. The training set contains 3,954 videos while the test set comprises 800 videos. XD-Violence covers various unusual types of events including abuse incidents, car accidents, explosions, fights, riots, and shootings. \textbf{UCF-Crime} dataset \cite{sultani2018real} is a large-scale collection of 1,900 videos captured by surveillance cameras in various indoor and outdoor scenarios. This dataset consists of 1,610 labeled training videos and 290 labeled test videos with a total duration of 217 hours. The dataset covers 13 types of anomalous events such as abuse, robbery, shootings and arson. \textbf{TAD} \cite{lv2021localizing} is a dataset for anomaly detection in traffic scenes, consisting of 400 training videos and 100 test videos, with a total of 25 hours of video footage. It covers seven types of real-world anomalies. \textbf{ShanghaiTech} consists of surveillance videos from different scenes on a campus \cite{zhang2016video}. The training set contains 237 videos while the testing set has 200 videos. 

\subsection{Implementation Details}
The dimension of the visual features $d_v$ extracted by CLIP(ViT-L/14) \cite{radford2021learning} is 768, while the dimension of semantic features $d_t$ is also 768. The prompt dictionary capacity $P$ is set to 30 for UCF-Crime, XD-Violence and TAD datasets, 25 for ShanghaiTech. The batch size $b$ is set to 64 for TAD dataset, and it is halved to 32 on the other three datasets. The number of snippets per video $L$ is set to 64 for all datasets. \hl{The feature operation $\theta$ is set as, a) concatenation for UCF-Crime, b) addition for the other three datasets.} The hyper-parameters $\alpha=1$, $\beta = 0.1$, $\gamma = 0.001$ and $\tau = 1$ are consistent across all datasets. The Adam optimizer is utilized with a learning rate of 0.001 and weight decay of 0.005 during the training process.

\begin{table}[]
\renewcommand\arraystretch{1.1}
\caption{Performance comparison of state-of-the-art methods on TAD (AUC\%) and ShanghaiTech (AUC\%). \textbf{Bold} and \underline{underline} indicate the best and second-best results.}
    \label{tab:tadsota}
\centering
\begin{tabular}{c|ccc|c|c}
    \toprule
    \toprule
    \multirow{2}{*}{\rotatebox{90}{Type}} & \multirow{2}{*}{Source} & \multirow{2}{*}{Method}& \multirow{2}{*}{Feat.} & TAD & ST\\ \cmidrule{5-6}
 & & && AUC &AUC \\
    \midrule
    \midrule
    \multirow{2}[1]{*}{\rotatebox{90}{Semi}}&ICCV 17' & Luo \textit{et al.} \cite{luo2017remembering} &-&57.9 &-\\
    & CVPR 18'& Liu \textit{et al.} \cite{liu2018future}& -&69.1 &72.8\\ \midrule
    \multirow{8}[1]{*}{\rotatebox{90}{Weakly}} & CVPR 21'& MIST \cite{feng2021mist}&UNet&89.2 &94.8\\
    & ICCV 21' & RTFM \cite{tian2021weakly}&I3D&89.6 &97.2\\
    & TIP 21' & WSAL \cite{lv2021localizing} &I3D& 89.6 &-\\
     & CVPR 23' & ECUPL \cite{zhang2023exploiting} &I3D&91.6 &-\\
     &CVPR 23' & CMRL \cite{cho2023look}&I3D& -&\textbf{97.6}\\
     &CVPR 23' & TEVAD \cite{chen2023tevad}&CLIP& 92.3&97.3\\
     & CVPR 23'&UMIL \cite{lv2023unbiased} &XCLIP& \underline{92.9} &96.8\\
     & \textbf{ours} & \textbf{LAP}&\textbf{CLIP}& \textbf{94.4}& \underline{97.4}\\
    \bottomrule
    \bottomrule
    \end{tabular}
\end{table}

\begin{figure*}[]
\centering
\includegraphics[width=1\textwidth]{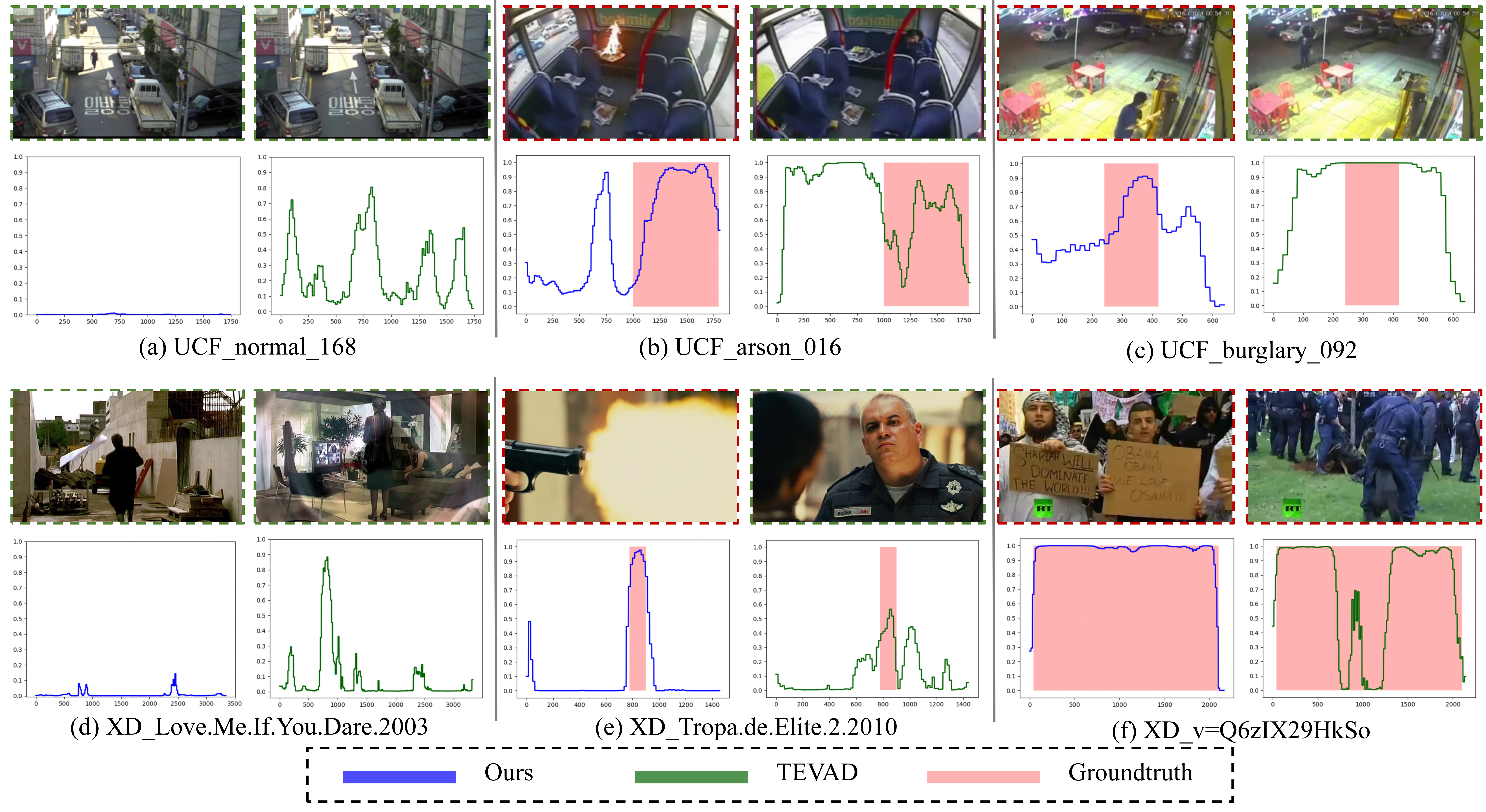}
\caption{Qualitative comparisons of TEVAD \cite{chen2023tevad} and our method on both UCF-Crime (UCF) and XD-Violence (XD). The ground truth of anomalous events is represented by light red regions.}
\label{fig:qualitative}
\end{figure*}

\subsection{Comparison Results}
\textbf{Quantitative analysis.} The comparisons between our LAP model and other state-of-the-art (SOTA) WS-VAD models on the XD-Violence and UCF-Crime datasets are presented in Table \ref{tab:xdsota}, and Table \ref{tab:tadsota} shows the comparisons on the TAD and Shanghaitech datasets.  It can be seen that the proposed model outperforms almost all the other methods in all datasets. 

Specifically, our model achieves the highest AP of 86.5\% on the XD-Violence dataset, outperforming the second best method VadCLIP \cite{wu2023vadclip} by 2.0\%, which also combines RGB and text data. Unlike the single-word description, i.e. class labels, used in VadCLIP, the event-level descriptions in our LAP model can provide much richer information, leading to a better understanding of the anomalies.
\hl{Another ClIP-based method (CLIP-TSA {\cite{joo2023clip}}) leverages the visual features from CLIP (VIT/B),} while a transformer is employed to enhance its features. However, due to the lack of semantic guidance, its AP falls 4.3\% below the proposed LAP. 
 The performance of the other compared the methods are also limited by the absence of efficient anomaly definitions. 

In the UCF-Crime dataset, our LAP model achieves an AUC$_{all}$ of 88.9\%, surpassing the most recent methods by at least 0.9\%, including VadCLIP \cite{wu2023vadclip} (88.0\%), CLIP-TSA \cite{joo2023clip} (87.6\%) and UMIL \cite{chen2023tevad} (86.7\%). Notably, the learnable prompts of VadCLIP are based on class labels, which leads to a video-level anomaly matching. While our concise descriptions of basic suspected anomaly events can effectively match the segment-level features. This difference results in a relatively high AUC$_{abn}$ of our LAP (73.0\%) comparing to VadCLIP (70.2\%). \hl{It is important to note that if we use more accurate text descriptions {\cite{yuan2024towards}} of each snippet, we can achieve a higher AUC$_{all}$ of 90.4\% and an AUC$_{abn}$ of 76.1\%. The results indicate that our model can effectively utilize the textual prompts of abnormal events for an accurate detection.}

Table \ref{tab:tadsota} shows the comparisons on the other two less challenging datasets. The AUC of our approach (94.4\%) is constantly higher than all SOTA methods compared \cite{zhang2023exploiting, lv2021localizing, lv2023unbiased} by a margin of 1.5\% to 4.8\% on TAD dataset. 
And our method achieves the second highest AUC (97.4\%) in ShanghaiTech, which is only 0.2\% lower than the best CMRL method \cite{cho2023look}. Noting that, if we switch our visual extractor from CLIP to I3D \cite{carreira2017quo} as the same as CMRL, the AUC will be boosted to 98.0\% (0.4\% higher CMRL). It indicates that the ShanghaiTech dataset is relatively less complex, while I3D \cite{carreira2017quo} is good enough. The details will be discussed in Section \ref{sec:dis}. For fair comparation, we re-implement TEVAD\cite{chen2023tevad} with visual features from CLIP. The AUC of our LAP exceeds TEVAD for 2.1\% in the TAD dataset and 0.1\% in the ShanghaiTech dataset. This minor performance gain on ShanghaiTech is due to the anomalies on campus being actually common activities such as riding, skating, and driving on the road, which are quite different from our suspected anomaly descriptions such as fighting, firing, or clashing.

Overall, these results highlight the superior performance of our LAP model compared to state-of-the-art methods on all four datasets in terms of both AP and AUC metrics. For fair comparison, the UMIL \cite{lv2023unbiased}, TEVAD \cite{chen2023tevad}, CLIP-TSA \cite{joo2023clip}, VadCLIP \cite{wu2023vadclip}, and the work by Wu \textit{et al.} \cite{wu2023open} are based on the same feature extractor (CLIP) as our LAP.

\textbf{Qualitative analysis.} To further demonstrate the effectiveness of our method, the qualitative comparisons between our approach (LAP) and the TEVAD SOTA method \cite{chen2023tevad} are visualized in Figure \ref{fig:qualitative}. The normal and abnormal frames of videos from the UCF-Crime and XD-Violence datasets are presented along with their corresponding frame-level anomaly scores, while green and red dashed rectangles indicate normal and abnormal ones, respectively.
As shown in the figures, our method not only outperforms TEVAD \cite{chen2023tevad} in terms of anomaly detection ability but also reduces false alarms on normal parts.

\begin{figure}[]
\centering
\includegraphics[width=.9\linewidth]{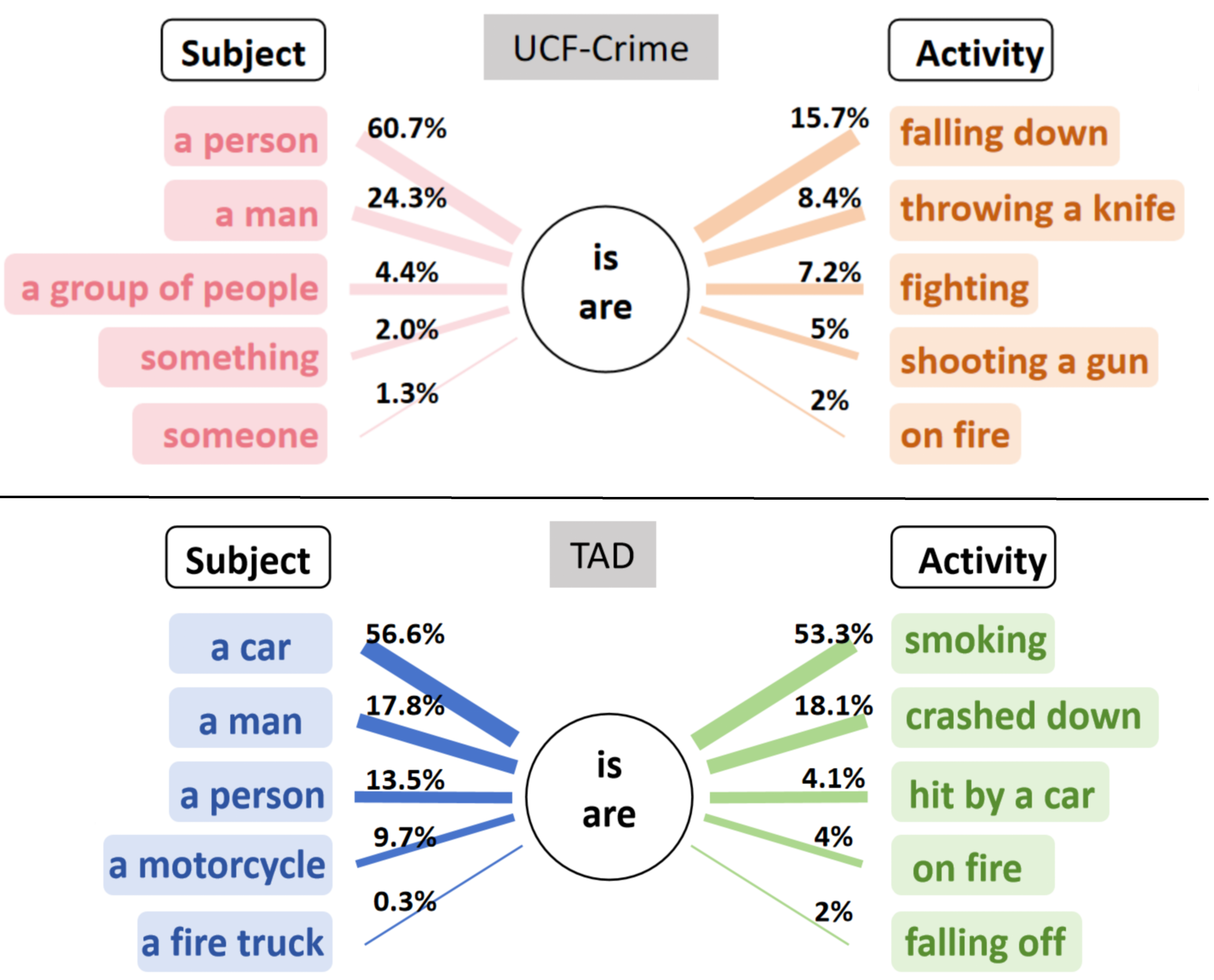}
\caption{The distribution of matched suspected anomalies in the UCF-Crime (upper) and TAD (lower) datasets.}
\label{fig:distribution}
\vspace{2mm}
\end{figure}

Our prompt dictionary contains event descriptions for various conditions. For the UCF-Crime social dataset and the TAD traffic dataset, Figure \ref{fig:distribution} illustrates the distributions of matched suspected anomalies in the anomaly vector $\textbf{c}$, showcasing how our prompt dictionary operates. Since the majority of anomalies in UCF-Crime are linked to human behavior like fights, robbery and violence, the predominant subject is ``person” and the most frequent activities include falling and using weapons. While in the other circumstance, the Traffic Anomaly Dataset (TAD) is consists of anomalies caused by traffic accidents. As expected, ``car” is the main subject, and the most common activities involve smoking and crashes. It indicates the effectiveness of our proposed event prompts. 

\subsection{Ablation Studies}
\label{sec:ablation}
\textbf{Components.} The proposed prompt-related components, i.e. feature synthesis (FS), multi-prompt learning (MPL) and pseudo anomaly labeling (PAL) are the keys to our superior performance in VAD. The ablation results of these three components on three datasets are shown in Table \ref{tab:componentablation}. The baseline module is a visual-only branch MIL-based network with a CLIP feature extractor \cite{radford2021learning}. By cooperating text branch for feature synthesis, our model achieves 1.2\%, 1.8\% and 0.4\% AUC$_{all}$ improvement, respectively, which shows the efficiency of semantic information. The MPL module can also improve the AUC$_{all}$ for all datasets by 0.3\%, 0.4\% and 0.1\%, while the AP on XD-Violence and AUC$_{abn}$ on UCF-Crime are also boosted by 0.9\% and 0.3\%. Further incorporating the PAL module yields better results. It outperforms the baseline by 1.9\% in AUC$_{all}$ and 6.0\% in AUC$_{abn}$ on UCF-Crime, 0.7\% in AUC$_{all}$ on TAD, as well as 2.4\% in AUC$_{all}$ and 5.2\% in AP on XD-Violence.

\begin{figure}[]
\centering
\captionsetup[subfigure]{font=footnotesize}
\subfloat[\label{fig:a} prompt format]{
\includegraphics[width=0.5\linewidth]{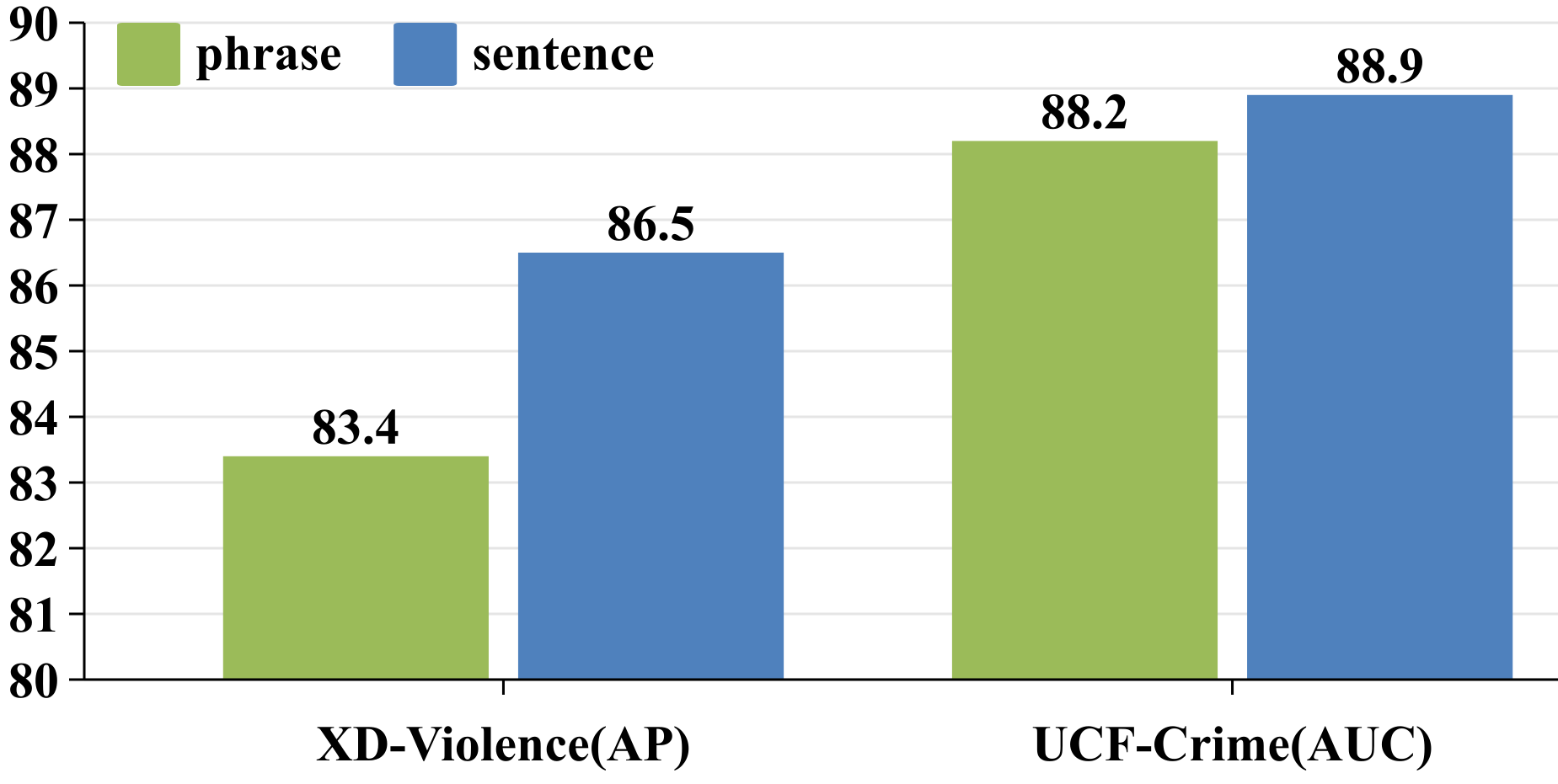}}
\subfloat[\label{fig:b} pseudo anomaly threshold]{	\includegraphics[width=0.5\linewidth]{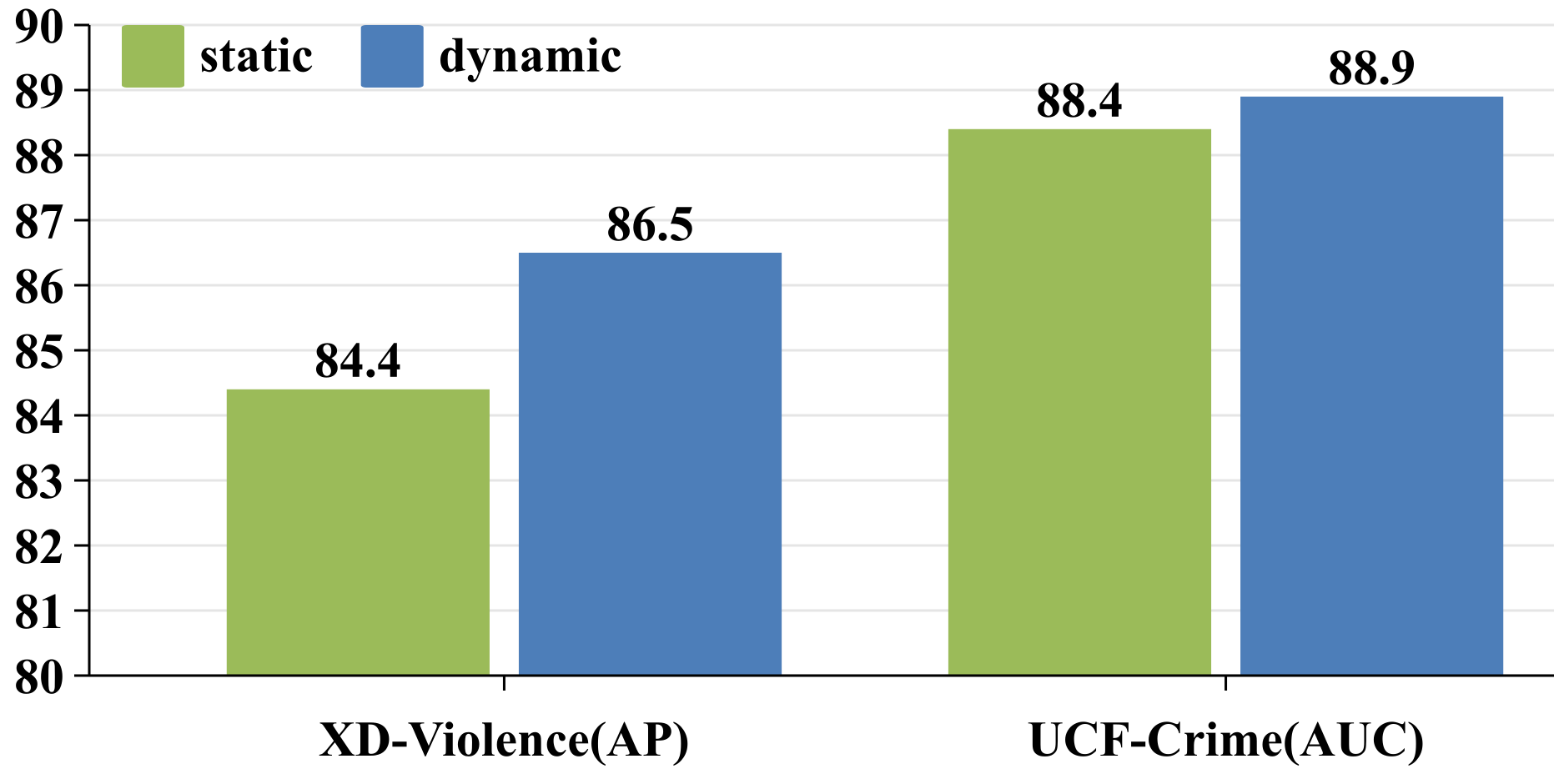}}
\caption{The ablation studies of the prompt format and pseudo anomaly threshold.}
\label{fig:prompt_ablation}
\vspace{2mm}
\end{figure}

\textbf{Prompt format.}
The prompt format plays an important role in the proposed LAP model. Thus, two different formats are tested in this experiment. One is organized by anomaly phrases, such as "falling down" or "on fire". The other contains complete anomaly sentences, like "someone is doing something to whom" or "something is what". As shown in Figure \ref{fig:a}, the sentence-based prompt dictionary outperforms the phrase-based one by 3.1\% on XD-Violence and 0.7\% on UCF-Crime, respectively. It suggests that prompts containing richer information are more helpful in identifying suspected anomalies.

\textbf{Pseudo anomaly threshold.} As required by the PAL module, pseudo labels are determined according to the threshold $\mathcal{G}_{h}$. The dynamic threshold given in Eq. \ref{eqn:thred} is used in previous experiments, which is based on the distribution of the data in the current batch. The hyper-parameter $\tau$ will determine the number for pseudo anomalies. When it is set to 0.5, 1.0 and 2.0, the AUC results on UCF-Crime are 88.05\%, 88.90\%, and 88.21\%, correspondingly. Another static threshold strategy is also compared in this test, while $\mathcal{G}_{h}$ is set to 0.5 as prior knowledge. As shown in Figure \ref{fig:b}, the dynamic threshold is better than the static one, whose AP and AUC are 2.1\% and 0.5\% higher on XD-Violence and UCF-Crime datasets.

\begin{table}[]
\caption{Ablation study of proposed modules. The default settings of all experiments are marked in gray color.}
  \label{tab:componentablation}%
  \centering
\renewcommand\arraystretch{1.1}
    \begin{tabular}{cccc|cc|cc|c}
    \toprule
    \toprule
    \multirow{3}{*}{\rotatebox{90}{Baseline\enspace}} &\multirow{2}{*}{\rotatebox{90}{ FS\,\enspace\enspace\enspace\enspace\enspace}}&\multirow{2}{*}{\rotatebox{90}{ MPL\,\,\enspace\enspace\enspace}}&\multirow{2}{*}{\rotatebox{90}{PAL\,\,\,\enspace\enspace\enspace}}& \multicolumn{2}{c|}{UCF-Crime} & \multicolumn{2}{c|}{XD-Violence}& TAD\\
    \cmidrule{5-9}
     && & & \multicolumn{2}{c|}{AUC}&AUC&AP&AUC \\
     &&&&${all}$&${abn}$&${all}$&${all}$&${all}$\\
    \midrule
    \midrule
    \checkmark & & & &87.0&67.0&93.2&81.3&93.7\\
    \midrule
    \checkmark  &\checkmark& & & 88.2& 70.4&95.0 &84.1 &94.1\\
    \midrule
    \checkmark  &\checkmark& \checkmark &  & 88.5& 70.7&95.4 &85.0 &94.2\\
    \midrule
    \checkmark  &\checkmark& \checkmark & \checkmark & \cellcolor{lightgray}\textbf{88.9} & \cellcolor{lightgray}\textbf{73.0} & \cellcolor{lightgray}\textbf{95.6}& \cellcolor{lightgray}\textbf{86.5}&\cellcolor{lightgray}\textbf{94.4}\\
    \bottomrule
    \bottomrule
    \end{tabular}
\end{table}

\begin{table}[]
    \caption{Comparisons of the AUC (\%) for open-set VAD on UCF-Crime. The numbers in braces are the amount of videos.}
    \label{tab:openset}
  \centering
  \vspace{2mm}
\renewcommand\arraystretch{1.1}
    \begin{tabular}{c|cccc}
    \toprule
    \toprule
    Open  & \multirow{2}{*}{No} & Explo- & RoadAcci- & Shoplif-\\
Category& &sion (21) & dents (23)& ting (21)\\
    \midrule
    \midrule
    RTFM \cite{tian2021weakly}& 84.3 & 83.6(-0.7) & 82.1(-2.2) & 83.4(-0.9)\\
    MLAD \cite{zhang2022weakly}& 85.4 & 84.3(-1.1) & 83.2(-2.2) & 84.5(-0.9)\\
    TEVAD \cite{chen2023tevad}& 84.9 & 83.7(-1.2) & 81.0(-3.9) & 83.1(-1.8)\\
    Ours & \textbf{88.9} & \textbf{88.1(-0.8)} & \textbf{87.0(-1.9)} &\textbf{88.4(-0.5)}\\
    \bottomrule
    \bottomrule
    \end{tabular}
\end{table}

\subsection{Discussions}
\label{sec:dis}
\begin{figure}[]
\centering
\includegraphics[width=1\linewidth]{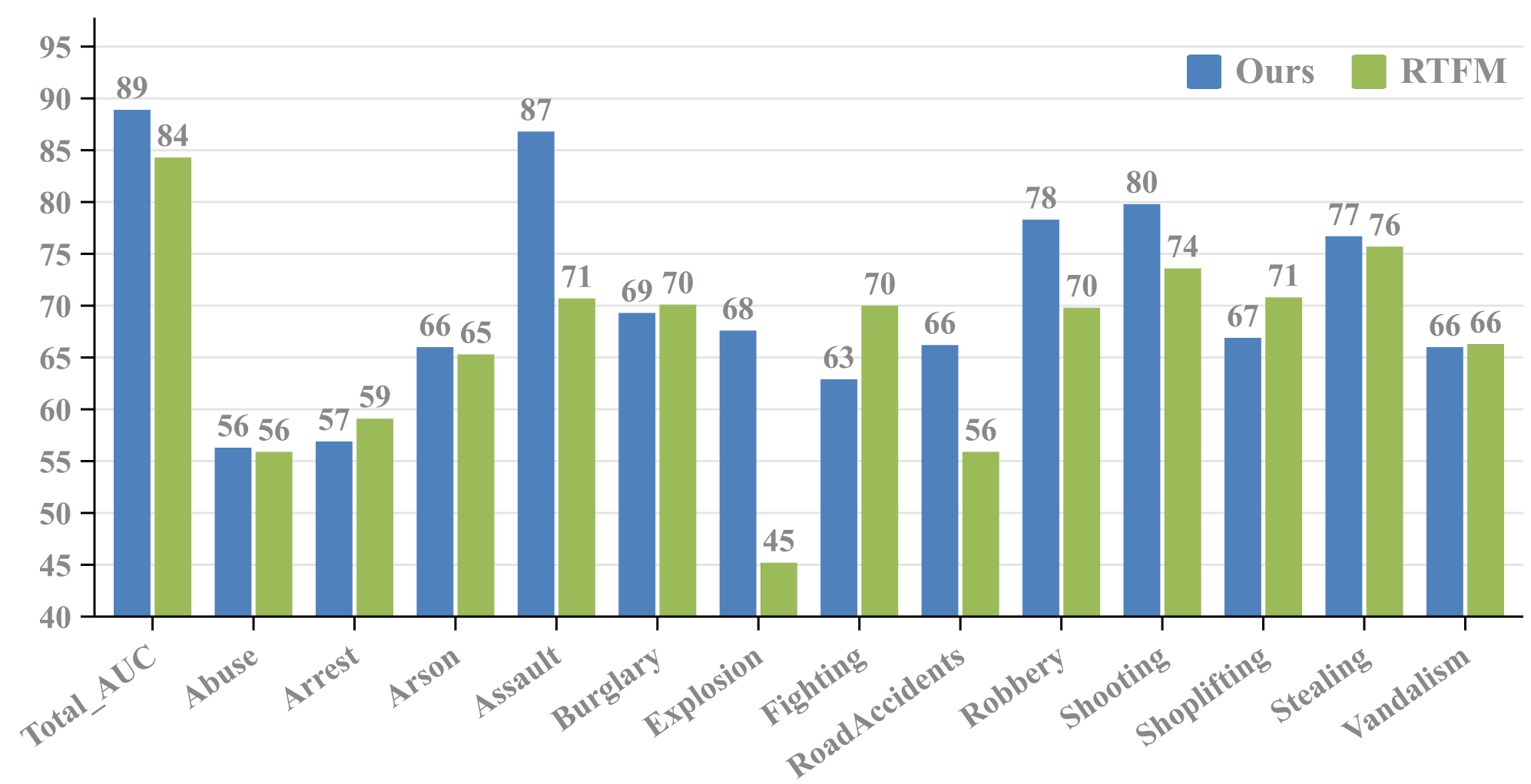}
\caption{Comparison of class-wise AUC (\%) on UCF-Crime dataset with RTFM \cite{tian2021weakly}.}
\label{fig:classauc}
\vspace{2mm}
\end{figure}

\textbf{Class-wise AUC.} To demonstrate the detailed performance on specific abnormal events, the class AUC of our model is compared with RTFM \cite{tian2021weakly} in Figure \ref{fig:classauc}. It shows that the proposed LAP model outperforms RTFM in most categories, especially on ``Assault", ``Explosion", ``RoadAccidents" and ``Robbery". This can be attributed to the effective use of our prompt dictionary to describe those anomalies including representative texts such as ``fire", ``knife" or ``accident". Combined with the MPL module, their synthetic features are more likely to be identified as abnormal ones. However, our model may be less effective in some cases if the action is subtle or difficult to describe, such as  ``Shoplifting" and ``Fighting" in Figure \ref{fig:classauc}.

\textbf{Open set VAD.}
In practical applications, it is impossible to collect or define all possible anomalies in advance. Hence, it is crucial to examine the robustness of anomaly detection models when confronted with open abnormal categories in real-world scenarios.
Following the protocol of open set VAD in MLAD \cite{zhang2022weakly}, experiments are conducted on the top 3 largest anomaly categories from the UCF-Crime dataset, namely ``Explosion", ``RoadAccidents" and ``Shoplifting". These categories are sequentially removed from the training set and treated as real open abnormal events.  
The comparisons with three SOTA models are presented in Table \ref{tab:openset}. It is obvious that the proposed LAP model outperforms RTFM \cite{tian2021weakly}, MLAD \cite{zhang2022weakly} and TEVAD \cite{chen2023tevad} in all three categories. It is worth noting that our method achieves minimal decreases in AUC values when compared to alternative approaches. This indicates that our method is more efficient in handling open abnormal event issues.

\begin{table}[]
  \caption{Cross-dataset experimental results on UCF-Crime (UCF) and XD-Violence (XD) benchmarks.}
    \label{tab:crossdataset}%
  \centering
  \vspace{2mm}
\renewcommand\arraystretch{1.1}
  \begin{tabular}{c|cc|cc}
    \toprule
    \toprule
Source & UCF & XD & XD & UCF \\ \midrule Target& \multicolumn{2}{c|}{UCF (AUC \%)}&\multicolumn{2}{c}{XD (AP \%)}\\
    \midrule
    \midrule
    RTFM \cite{tian2021weakly}& 84.3 & 68.6 \textcolor{gray}{(-15.7)} & 76.6 & 37.3 \textcolor{gray}{(-39.3)} \\
        CMRL \cite{cho2023look}& 86.1 & 69.9 \textcolor{gray}{(-16.2)} & 81.3 & 46.7 \textcolor{gray}{(-34.6)} \\
        Ours& \textbf{88.9}& \textbf{83.5}\textbf{ \textcolor{gray}{(-5.4)}}& \textbf{86.5}& \textbf{ 60.9\textcolor{gray}{(-25.6)}}\\
    \bottomrule
    \bottomrule
    \end{tabular}%
\end{table}%

\textbf{Cross-dataset performance.}
The categories of anomalies varies from different VAD datasets. For instance, the abnormal events in UCF-Crime dataset are collected from surveillance videos, which is quite different from the abnormal categories in XD-Violence developed from movies and online videos. Thus, it will become a challenging transfer learning task, if the model is trained and inferred on different datasets. However, it is actually what will happen in real-world anomaly detection applications. To evaluate the generalization and zero-shot abilities of our proposed method, another set of experiments using different sources of training and inference videos is conducted. 
Compared with RTFM \cite{tian2021weakly} and CMRL \cite{cho2023look}, our model explains the definition of anomalies with their descriptions using the prompt dictionary and multi-prompt learning scheme. Such a new paradigm of utilizing the semantic information leads to an extraordinary cross-dataset performance as shown in Table \ref{tab:crossdataset}. The performance degradation of the proposed model is only one-third of the ones of RTFM and CMRL, when it is trained on XD-Violence and tested on ShanghaiTech. It indicates that our method is much less sensitive to variations in the data domain, which is important for practical applications. 

\begin{table}[]
\caption{Performance of MPL and PAL embedded RTFM \cite{tian2021weakly} on ShanghaiTech.}
\label{tab:r1}
\renewcommand\arraystretch{1.1}
  \centering
  \begin{tabular}{c|c|c|c|c}
\toprule \toprule
        Method (ST) & AUC$_{all}$&AUC$_{abn}$  &FAR$_{all}$ &FAR$_{abn}$\\
    \midrule
    RTFM & 97.2 &64.3 &0.06 &0.86\\
 RTFM+MPL  & 97.6& 72.2&0.06 &0.71\\
 RTFM+PAL  &97.5 &73.9 &\textbf{0.03} &\textbf{0.44}\\
    RTFM+Both & \textbf{98.0}  &\textbf{75.6} &0.04 &0.58\\
    \bottomrule \bottomrule
     \end{tabular}%
\vspace{2mm}
\end{table}

\textbf{Plug and play.}
To further explore the potential of our method, the proposed MPL and PAL modules are embedded into the representative WS-VAD work RTFM \cite{tian2021weakly} on ShanghaiTech. For a fair comparison, I3D \cite{carreira2017quo}, instead of CLIP \cite{radford2021learning}, is used as the feature extractor, while all experimental settings were kept the same as in RTFM paper. As shown in Table \ref{tab:r1}, the reimplemented frameworks generally exhibited better performance. By incorporating either MPL or PAL alone, enhancements of 0.4\% and 0.3\% can be achieved on AUC$_{all}$, whereas more significant enhancements of 7.9\% and 9.6\% can be observed on AUC$_{abn}$. The efficacy of MPL and PAL is demonstrated by their ability to improve the performance of the conventional WSAD framework. Through the collaboration of MPL and PAL, LAP integrated RTFM demonstrates superior AUC and reduced FAR compared to its original version, showing significant enhancements (0.8\%, 0.02\%) on the ShanghaiTech dataset. It is worth noting that the reimplemented RTFM model (98. 0\%) could even surpass the latest SOTA model CMRL \cite{cho2023look} by 0.4\%. It indicates that more frameworks may benefit from our prompt-related modules, which are plug-and-play.

\section{Conclusion}

In this study, we presented the LAP model, a straightforward yet effective method for WS-VAD. Specifically, the synthesized visual-semantic features have been employed for better feature representation. The multi-prompt learning strategy has shown its capability to guide the learning of suspected anomalies with a prompt dictionary. Additionally, the pseudo anomaly labels generated by the anomaly similarity between the prompts and video captions are useful to enhance the VAD performance. Extensive experiments have demonstrated the effectiveness of our model. We hope that our work will inspire further exploration of defining and learning anomalies from natural languages.

\bibliographystyle{IEEEtran}
\bibliography{IEEEabrv,reference}

\end{document}